\documentclass{article}

\usepackage[utf8]{inputenc} 
\usepackage{amsmath}        
\usepackage{amssymb}        
\usepackage{graphicx}       
\usepackage[hidelinks]{hyperref} 

\usepackage{booktabs}       
\usepackage{geometry}       
\usepackage{float}          

\title{The Domain Mixed Unit: A New Neural Arithmetic Layer}
\author{Paul Curry}
\date{\today}

\begin{document}

\maketitle

\begin{abstract}
The Domain Mixed Unit (DMU) is a new neural arithmetic unit that learns a single parameter gate G that mixes a state between log-space and linear-space representations while performing either addition (DMU\_add) or subtraction (DMU\_sub) in said space. These are the two initializations proposed for the DMU: one covering addition and multiplication, and another covering subtraction and division. The DMU achieves state-of-the-art performance on the NALM Benchmark, a dataset designed to test the ability of neural arithmetic units to generalize arithmetic operations, specifically performing with the highest percentage solved over all seeds on multiplication and division. The DMU will be submitted as a pull request to the open-source NALM benchmark, and its code is currently available on GitHub under the forked repository: \url{https://github.com/marict/nalm-benchmark}.
\end{abstract}

\section{Neural Arithmetic Units}
Neural Arithmetic Units (NAUs) are specialized sub-units or networks designed to interpretably represent arithmetic operations while maintaining differentiability, allowing gradients to flow through them during training. These units can be integrated into larger neural architectures to provide explicit arithmetic capabilities.

\section{Justification}
The hypothesis underlying this work is that the inclusion of Neural Arithmetic Units (NAUs) will improve a model's ability to perform arithmetic operations. It has been shown that small models often struggle with arithmetic \cite{lake2018, suzgun2019, trask2018}. Research from Anthropic further indicates that large models form neural circuits capable of arithmetic, but in convoluted ways that fail to generalize with precision \cite{anthropic2023}. 

By providing models with explicit arithmetic layers that have stronger inductive biases toward interpretable arithmetic operations, we can potentially guide the model to utilize these more principled structures instead of the naturally occurring but less reliable ones. The idea is to place these layers inside larger models (either frozen, randomly initialized and trained, or initialized to specific operations and trained further), such as language models, to enhance their ability to perform simple arithmetic.

This approach offers several advantages: (1) improved arithmetic performance without requiring external tool calls, (2) reduced token usage in reasoning chains, and (3) enhanced interpretability of the model's arithmetic processing. If given interpretable networks that can fully represent arithmetic operations to high precision, these inductive bias circuits may be used instead of the naturally occurring structures, resulting in higher performance. The DMU represents an attempt to create such an inductive bias that can be seamlessly integrated into existing neural architectures while maintaining differentiability and end-to-end trainability.

\section{Prior Work}
Several neural arithmetic layers have been proposed and benchmarked on the NALM dataset. For a detailed description of each variant, see A Primer for Neural Arithmetic Logic Modules (Mistry et al., 2022) \cite{mistry2022primerneuralarithmeticlogic}.

\subsection{Neural Arithmetic Logic Unit (NALU)}
The Neural Arithmetic Logic Unit (NALU) (Trask et al., 2018) \cite{trask2018} introduced inductive biases for learning addition, subtraction, multiplication, and division with extrapolation capabilities. Its architecture combines an additive sub-unit (NAC+) and a multiplicative sub-unit (NAC•), with a gating mechanism selecting between them. 

\subsection{Improved NALU (iNALU)}
The iNALU was developed to fix shortcomings of the original NALU. Improvements include independent weights for additive and multiplicative paths, a mechanism to correctly handle negative signs in multiplication, and reinitializing weights during training if convergence stalls as well as regularization and clamping. The clamping mechanism in the DMU is similar to that in the work.

\subsection{Neural Addition Unit (NAU) and Neural Multiplication Unit (NMU)}
The NAU handles addition and subtraction with weights constrained to $\{-1, 0, 1\}$, while the NMU handles multiplication with weights constrained to $\{0, 1\}$. Both enforce sparsity to keep weights interpretable and avoid the instabilities of NALU, making them cleaner, more robust building blocks. Complex arithmetic functions are composed by stacking NAUs and NMUs rather than using a single gated unit.

\subsection{Neural Power Unit (NPU) and RealNPU}
The NPU extends arithmetic modules to handle division and arbitrary powers by using complex log transformations with real and imaginary weight matrices. Unlike NAU/NMU, weights need not be discrete (e.g., 0.5 for square roots). A relevance gate ensures only selected inputs contribute, preventing collapse to zero. A simplified variant, the RealNPU, uses only real values. Regularization is applied via a scaled L1 penalty rather than discrete weight constraints, preserving flexibility.

\subsection{Golden Ratio NALU (G-NALU)}
The G-NALU is a variant of the NALU that replaces the natural base $e$ in its activation functions with the golden ratio ($\phi \approx 1.618$). This change is intended to produce smoother gradients, potentially stabilizing training.

\section{NALM Dataset Task}
The Neural Arithmetic Logic Modules (NALM) benchmark \cite{madsen2019measuring} is designed to evaluate the ability of neural arithmetic units to generalize arithmetic operations beyond their training distributions. The task structure is straightforward: given an input vector $\mathbf{x} \in \mathbb{R}^2$, the model must produce a scalar output $y \in \mathbb{R}^1$ where $y$ represents the result of applying a specific arithmetic operation to the components of $\mathbf{x}$.

Crucially, only one arithmetic operation is benchmarked at a time. For a given evaluation, the model is trained and tested exclusively on one of the four fundamental operations: addition ($x_1 + x_2$), subtraction ($x_1 - x_2$), multiplication ($x_1 \times x_2$), or division ($x_1 \div x_2$). This allows for focused evaluation of how well each neural arithmetic unit can learn and generalize the specific mathematical relationship.

Generalization performance is evaluated using a weight perturbation method where optimal solution weights are perturbed by $\epsilon = 1 \times 10^{-5}$, and extrapolation loss is computed over 1 million examples, resulting in a specific threshold for each neural arithmetic unit, range, and operation combination. The threshold values used for the DMU evaluation are provided in the appendix, with representative values ranging from $10^{-9}$ for the most precise operations to $10^{-1}$ for the most challenging extrapolation scenarios.

\begin{figure}[H]
\centering
\begin{equation*}
\boxed{\begin{array}{c} \mathbf{x} \in \mathbb{R}^2 \\ \text{(Input)} \end{array}} \xrightarrow{\text{Operation}} \boxed{\begin{array}{c} o \in \{+, -, \times, \div\} \\ \text{(Function)} \end{array}} \xrightarrow{\text{Apply}} \boxed{\begin{array}{c} y \in \mathbb{R}^1 \\ \text{(Ground Truth)} \end{array}}
\end{equation*}

\vspace{0.5cm}

\textbf{Examples:}
\begin{align*}
[2.5, 1.3] &\xrightarrow{\text{Addition}} 3.8 \\
[2.5, 1.3] &\xrightarrow{\text{Multiplication}} 3.25
\end{align*}

\caption{NALM Dataset Task Structure: The task maps a 2D input vector to a 1D output through a single arithmetic operation. Only one operation is evaluated per experiment.}
\label{fig:nalm_task}
\end{figure}

\section{Domain Mixed Unit}
The proposed architecture in this paper, the Domain Mixed Unit (DMU), borrows several elements from prior work and expands upon the design.

\subsection{Structure}
The core computation of the DMU begins by decomposing the input vector $\mathbf{x} \in \mathbb{R}^2$ into its magnitude $|\mathbf{x}|$ and sign $\text{sign}(\mathbf{x})$. The sign information is processed on a separate path for two key reasons:
\begin{enumerate}
    \item It allows the use of log-space operations while preserving crucial sign information.
    \item It enables the value mixture computation to be performed entirely in  log-space for numerical stability.
\end{enumerate}

\subsubsection*{Linear and Log-Space Computations}
First, the unit performs the selected operation in both the linear and space on the input magnitudes using an operand selector matrix $\mathbf{O}$. This yields a linear result, $y_{\text{lin}}$, and a log-space result, $y_{\text{log}}$. The log-space operation includes clamping to prevent numerical instabilities.
\begin{align*}
    y_{\text{lin}} &= \mathbf{O} \cdot |\mathbf{x}| && \text{(Linear Space operation)} \\
    y_{\text{log}} &= \mathbf{O} \cdot \log(\text{clamp}(|\mathbf{x}|, \text{min}=\epsilon_{\text{mag}})) && \text{(Log Space operation)}
\end{align*}

\subsubsection*{Sign Calculation}
Corresponding signs for each path are then computed. For the linear path, the sign $s_{\text{lin}}$ is a scaled hyperbolic tangent. For the log path, a cosine function is used to map the number of negative inputs to the correct output sign for multiplication or division, yielding $s_{\text{log}}$.
\begin{align*}
    s_{\text{lin}} &= \tanh(y_{\text{lin}} / T) && \text{(Linear path sign)} \\
    s_{\text{log}} &= \cos\left(\pi \cdot \sum \frac{1 - \text{sign}(\mathbf{x})}{2}\right) && \text{(Log path sign)}
\end{align*}
The behavior of the log path sign calculation is illustrated in Table \ref{tab:sign_calc}.

\begin{table}[!htbp]
\centering
\caption{Example of Log Space Sign Calculation}
\label{tab:sign_calc}
\begin{tabular}{ccccc}
\toprule
\textbf{Signs} & \textbf{n} & \textbf{m} & \textbf{cos} & \textbf{result} \\
\midrule
(1,1) & (0,0) & 0 & $\cos(0)$ & 1 \\
(1,-1) & (0,1) & 1 & $\cos(\pi)$ & -1 \\
(-1,-1) & (1,1) & 2 & $\cos(2\pi)$ & 1 \\
\bottomrule
\end{tabular}
\end{table}

\subsubsection*{Domain Mixing}
Next, a learned gating vector $\mathbf{G} = [G_{\text{lin}}, G_{\text{log}}]$ mixes the results from the two domains. The signs are mixed directly to produce the final mixed sign, $s_{\text{mix}}$:
\begin{equation*}
    s_{\text{mix}} = G_{\text{lin}} \cdot s_{\text{lin}} + G_{\text{log}} \cdot s_{\text{log}}
\end{equation*}

The magnitudes are mixed in log space to prevent value blowouts and gradient explosions. To achieve this, the result of the linear path, $y_{\text{lin}}$, is first mapped to a non-negative value using a smooth absolute value approximation and then transformed into log space with clamping:
\begin{align*}
    |y_{\text{lin}}|_{\epsilon} &= \sqrt{y_{\text{lin}}^2 + 10^{-8}} && \text{(Smooth absolute value)} \\
    \hat{y}_{\text{lin}} &= \log(\text{clamp}(|y_{\text{lin}}|_{\epsilon}, \text{min}=\epsilon_{\text{mag}})) && \text{(Map linear result to log space)} \\
    \tilde{y}_{\text{log}} &= \text{clamp}(y_{\text{log}}, \text{min}=-L, \text{max}=L) && \text{(Clamp log values)}
\end{align*}
The final mixed magnitude, $M_{\text{log}}$, is a weighted sum of the two paths, computed in log space with final clamping:
\begin{align*}
    M_{\text{log}}' &= G_{\text{lin}} \cdot \hat{y}_{\text{lin}} + G_{\text{log}} \cdot \tilde{y}_{\text{log}} \\
    M_{\text{log}} &= \text{clamp}(M_{\text{log}}', \text{min}=-L, \text{max}=L) && \text{(Final magnitude clamp)}
\end{align*}

\subsubsection*{Final Output}
Finally, the mixed magnitude is converted back to linear space via exponentiation to get $M_{\text{final}}$, and combined with the mixed sign $s_{\text{mix}}$ to produce the final output, $y_{\text{final}}$:
\begin{align*}
    M_{\text{final}} &= e^{M_{\text{log}}} \\
    y_{\text{final}} &= s_{\text{mix}} \cdot M_{\text{final}}
\end{align*}

\begin{verbatim}
Algorithm: Domain Mixed Unit (DMU) Forward Pass

Function DMU(x, O, G, T, epsilon, mag_min, log_lim):
    // Decompose input vector x in R^2
    x_mag = |x|
    x_sign = sign(x)
    
    // Calculate results in linear and log space for magnitudes
    r_lin = O * x_mag
    r_log = O * log(clamp(x_mag, min=mag_min))  // Clamp before log
    
    // Calculate corresponding signs for both spaces
    sign_lin = tanh(r_lin / T)
    n = 0.5 * (1.0 - x_sign)
    m = sumn
    sign_log = cos(pi * m)
    
    // Mix the signs using the gate G = [G_lin, G_log]
    V_sign = G_lin * sign_lin + G_log * sign_log
    
    // Mix the magnitudes in log space
    mag_lin = sqrt(r_lin^2 + epsilon)  // Smooth absolute value
    l_lin = log(clamp(mag_lin, min=mag_min))  // Clamp before log
    l_log = clamp(r_log, min=-log_lim, max=log_lim)  // Clamp log values
    m_log = G_log * l_log + G_lin * l_lin
    m_log = clamp(m_log, min=-log_lim, max=log_lim)  // Final clamp
    
    // Convert final magnitude back to linear space
    V_mag = e^(m_log)
    
    // Combine final sign and magnitude
    final_value = V_sign * V_mag
    
    Return final_value
\end{verbatim}

\subsection{NALM Configuration}
In the simplified configuration used for evaluation on the NALM dataset, the operand selector $O \in \{1, -1\}^2$ is frozen to canonical patterns $[1,1]$ for addition (DMU\_add) and $[1, -1]$ for subtraction (DMU\_sub). With these fixed operand patterns, the gating parameter $G$ becomes the primary learnable component that determines how to mix linear- and log-domain computations for the two selected inputs.

\section{Results}
DMU\_add and DMU\_sub both perform optimally on the NALM Benchmark task for their respective operations. The NALM benchmarking task is a benchmark designed to measure the ability of neural arithmetic units to generalize arithmetic operations. Both cases of the DMU both completely extrapolate their arithmetic operations in a reasonable time given the threshold determined from perturbing the optimal weight value, which in this case are the only parameters, the two gate values. The DMU has no differing behavior between different seeds. The only difference required in training for the DMU to extrapolate on all tasks is to increase the learning rate from the default 1e-3 to 1e-2.

The following tables compare the performance of the DMU\_add and DMU\_sub models against prior work, with baseline results sourced from the NALM benchmark primer \cite{mistry2022primerneuralarithmeticlogic}.

\subsection{Statistical Analysis}
Comprehensive evaluation across 900 total experiments (25 seeds per 4 operations and 9 ranges) reveals the DMU's exceptional performance consistency with 100\% success rate across all operations.

\begin{table}[!htbp]
\centering
\caption{DMU Performance Analysis}
\label{tab:dmu_stats}
\small
\begin{tabular}{lccr}
\toprule
\textbf{Operation} & \textbf{Mean Conv. Time} & \textbf{Mean Sparsity} & \textbf{Mean Extrapolation Error} \\
\midrule
ADD & 2170 & 0.281 & $3.3 \times 10^{-6}$ \\
SUB & 7842 & 0.280 & $1.5 \times 10^{-6}$ \\
MUL & 5626 & 0.266 & $1.4 \times 10^{-5}$ \\
DIV & 1400 & 0.277 & $7.3 \times 10^{-8}$ \\
\bottomrule
\end{tabular}
\end{table}

The mean sparsity values indicate effective parameter utilization, while the extremely low extrapolation error values demonstrate precise convergence to target functions across all evaluation ranges.

\subsection{Operation Task Results}

\begin{table}[!htbp]
\centering
\caption{Addition Success Rates (\%) Comparison}
\label{tab:add_success}
\footnotesize
\begin{tabular}{l|ccccc}
\toprule
\textbf{Range} & \textbf{DMU\_add} & \textbf{NAU} & \textbf{iNALU} & \textbf{NALU} & \textbf{G-NALU} \\
\midrule
U[-0.2, -0.1) & \textbf{100\%} & 100\% & 100\% & 52\% & 24\% \\
U[-1.2, -1.1) & \textbf{100\%} & 100\% & 100\% & 40\% & 0\% \\
U[-20, -10) & \textbf{100\%} & 100\% & 100\% & 68\% & 8\% \\
U[-2, -1) & \textbf{100\%} & 100\% & 100\% & 64\% & 0\% \\
U[0.1, 0.2) & \textbf{100\%} & 100\% & 100\% & 16\% & 12\% \\
U[1.1, 1.2) & \textbf{100\%} & 100\% & 100\% & 76\% & 0\% \\
U[10, 20) & \textbf{100\%} & 100\% & 100\% & 20\% & 4\% \\
U[1, 2) & \textbf{100\%} & 100\% & 100\% & 80\% & 0\% \\
U[-2, 2) & \textbf{100\%} & 100\% & 100\% & 0\% & 0\% \\
\bottomrule
\end{tabular}
\end{table}
\begin{table}[!htbp]
\centering
\caption{Subtraction Success Rates (\%) Comparison}
\label{tab:sub_success}
\footnotesize
\begin{tabular}{l|ccccc}
\toprule
\textbf{Range} & \textbf{DMU\_sub} & \textbf{NAU} & \textbf{iNALU} & \textbf{NALU} & \textbf{G-NALU} \\
\midrule
U[-0.2, -0.1) & \textbf{100\%} & 100\% & 100\% & 0\% & 0\% \\
U[-1.2, -1.1) & \textbf{100\%} & 100\% & 40\% & 0\% & 0\% \\
U[-20, -10) & \textbf{100\%} & 100\% & 100\% & 20\% & 4\% \\
U[-2, -1) & \textbf{100\%} & 100\% & 100\% & 12\% & 0\% \\
U[0.1, 0.2) & \textbf{100\%} & 100\% & 100\% & 0\% & 0\% \\
U[1.1, 1.2) & \textbf{100\%} & 100\% & 56\% & 0\% & 0\% \\
U[10, 20) & \textbf{100\%} & 100\% & 100\% & 84\% & 20\% \\
U[1, 2) & \textbf{100\%} & 100\% & 100\% & 12\% & 0\% \\
U[-2, 2) & \textbf{100\%} & 100\% & 100\% & 0\% & 0\% \\
\bottomrule
\end{tabular}
\end{table}

\begin{table}[!htbp]
\centering
\caption{Multiplication Success Rates (\%) Comparison}
\label{tab:mul_success}
\footnotesize
\begin{tabular}{l|cccccc}
\toprule
\textbf{Range} & \textbf{DMU\_add} & \textbf{NMU} & \textbf{iNALU} & \textbf{RealNPU} & \textbf{NALU} & \textbf{G-NALU} \\
\midrule
U[-0.2, -0.1) & \textbf{100\%} & 100\% & 100\% & 0\% & 0\% & 0\% \\
U[-1.2, -1.1) & \textbf{100\%} & 68\% & 12\% & 0\% & 0\% & 0\% \\
U[-20, -10) & \textbf{100\%} & 100\% & 12\% & 0\% & 52\% & 16\% \\
U[-2, -1) & \textbf{100\%} & 80\% & 68\% & 0\% & 12\% & 0\% \\
U[0.1, 0.2) & \textbf{100\%} & 100\% & 4\% & 12\% & 60\% & 32\% \\
U[1.1, 1.2) & \textbf{100\%} & 100\% & 100\% & 28\% & 0\% & 0\% \\
U[10, 20) & \textbf{100\%} & 100\% & 100\% & 84\% & 84\% & 16\% \\
U[1, 2) & \textbf{100\%} & 100\% & 100\% & 100\% & 8\% & 0\% \\
U[-2, 2) & \textbf{100\%} & 100\% & 100\% & 8\% & 0\% & 0\% \\
\bottomrule
\end{tabular}
\end{table}

\begin{table}[!htbp]
\centering
\caption{Division Success Rates (\%) Comparison}
\label{tab:div_success}
\footnotesize
\begin{tabular}{l|cccccc}
\toprule
\textbf{Range} & \textbf{DMU\_sub} & \textbf{iNALU} & \textbf{RealNPU} & \textbf{NPU} & \textbf{NALU} & \textbf{NAC•} \\
\midrule
U[-0.2, -0.1) & \textbf{100\%} & 0\% & 32\% & 0\% & 0\% & 0\% \\
U[-1.2, -1.1) & \textbf{100\%} & 0\% & 12\% & 0\% & 0\% & 0\% \\
U[-20, -10) & \textbf{100\%} & 0\% & 4\% & 0\% & 0\% & 16\% \\
U[-2, -1) & \textbf{100\%} & 0\% & 84\% & 0\% & 0\% & 8\% \\
U[0.1, 0.2) & \textbf{100\%} & 100\% & 88\% & 88\% & 0\% & 0\% \\
U[1.1, 1.2) & \textbf{100\%} & 100\% & 16\% & 16\% & 0\% & 0\% \\
U[10, 20) & \textbf{100\%} & 20\% & 4\% & 4\% & 0\% & 16\% \\
U[1, 2) & \textbf{100\%} & 100\% & 100\% & 100\% & 0\% & 8\% \\
U[-2, 2) & \textbf{100\%} & 0\% & 0\% & 0\% & 0\% & 0\% \\
\bottomrule
\end{tabular}
\end{table}

\section{Loss Landscapes}
The exceptional performance of DMU\_add and DMU\_sub can be attributed to their operation in an optimal parameter space. When the operand selector $O$ is frozen to the correct canonical patterns and only the gating parameter $G$ is learned, the loss landscape forms a smooth, convex function that enables reliable convergence to the global optimum.

\begin{figure}[!htbp]
\centering
\includegraphics[width=0.8\textwidth]{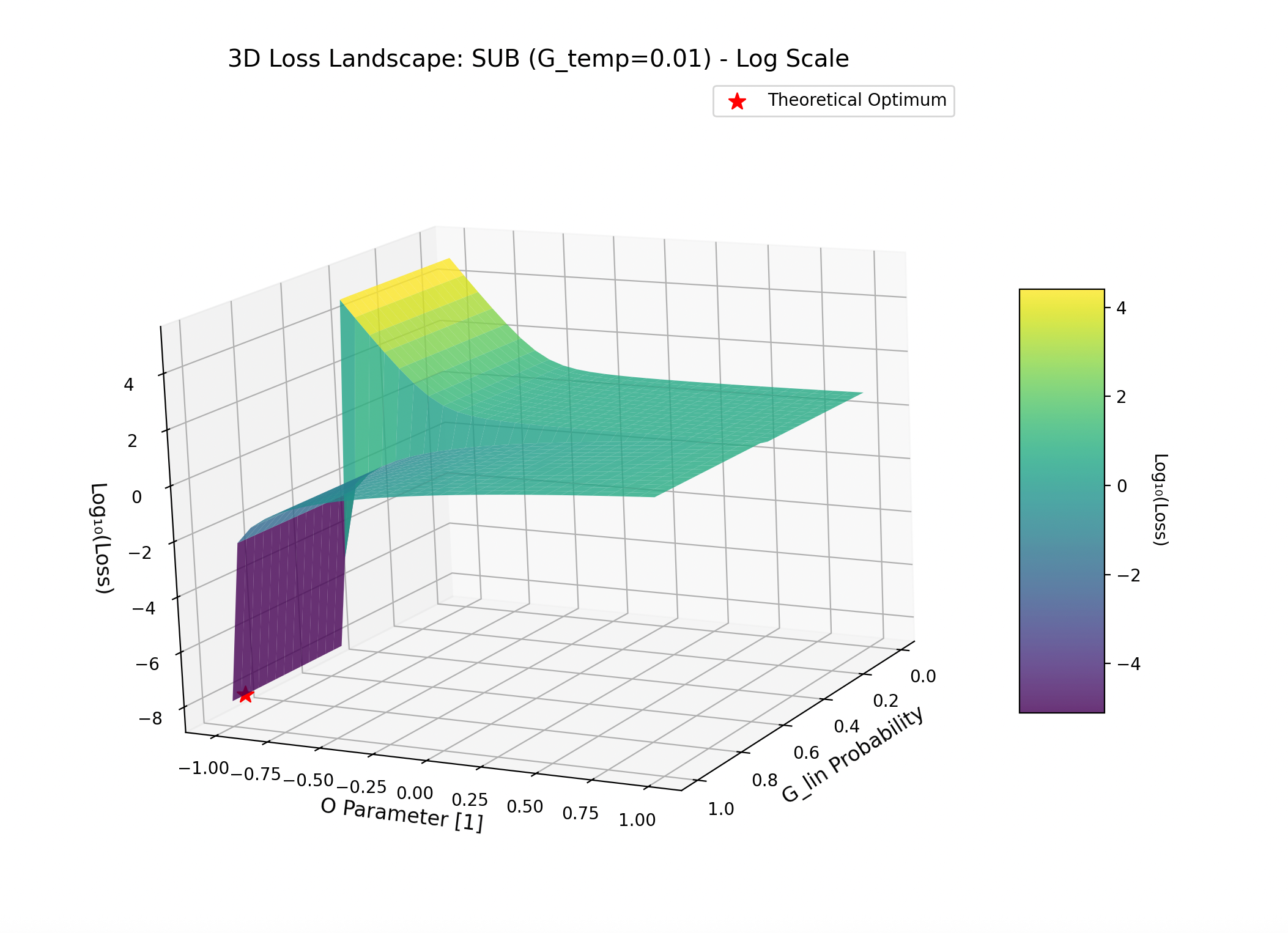}
\caption{Loss landscape for subtraction operation showing a problematic slope that pushes the unit toward an incorrect $O[1]$ value.}
\label{fig:sub_loss}
\end{figure}

However, when both $O$ and $G$ are simultaneously optimized, the loss landscape becomes significantly more complex, characterized by numerous saddle points that impede convergence. This phenomenon is particularly pronounced in subtraction tasks, where the loss surface exhibits slopes that push the model parameters toward incorrect values of $O[1]$, as illustrated in Figure \ref{fig:sub_loss}.

The presence of these saddle points reveals a fundamental challenge in designing universal neural arithmetic units. While the DMU achieves perfect performance on individual operations through careful parameter initialization and selective optimization, creating a single unit capable of learning all four arithmetic operations simultaneously requires sophisticated methods to smooth and resolve these problematic regions in the loss landscape.

If a Neural Arithmetic Unit is to be created that performs well on all operations without operation-specific initialization, it must incorporate mechanisms to navigate around or eliminate these saddle points, potentially through architectural innovations, specialized training procedures, or hybrid approaches that combine the benefits of both fixed and learnable components.

\section{Discussion}
The DMU provides a significant simplification over previous arithmetic variants, which may play a role in its ability to converge reliably. It is important to note that when compositionality is added to the DMU and/or the DMU has its O selector unfrozen, the ability to converge on each range and operation is greatly reduced, meaning that the G gate provides a smooth gradient descent whereas the introduction of other factors increases the presence of saddle points in the loss space. Finding a smooth way to align both the axes of the operation and space the operation is performed in is the key to creating an arithmetic unit that can generalize across all operations.

The temperature scalar of 0.1 on the G gate is required for the smallest negative range on multiplication n01 to converge, this is due to a saddle point being reached where a 20/80 lin/log mix approximates most multiplication operations. For smaller ranges likely an even tighter temperature scalar would be required to enforce discretization of G, causing instability at a certain point.

\section{Extensibility}
The DMU can be extended to express arbitrary directed acyclic graphs of depth N. The extension is as follows. $M = \text{num\_initial\_values} + \text{dag\_depth}$. O is unfrozen and extended to size $M \times N$. It is masked such that for row $n$, only $\text{num\_initial\_values} + n$ are non-zero. G is turned into a size $N$ vector, determining the space in which $O$ is applied in each step.
\begin{itemize}
    \item Predict $O$ and $G$ given the input, this can be done via a simple MLP head for each parameter or something more parameter dense.
    \item Start the dag execution:
    \begin{itemize}
        \item A working\_mag and working\_sign vector are initialized as size $M$.
        \item The first half being for the initial values and the second half being for the immediate values.
        \item For each dag step $n$:
        \begin{itemize}
            \item Apply log/lin calculation and mixing step for the current working\_mag and given $O$.
            \item Append the final value and sign to the $\text{num\_initial\_values} + n$ slot of the working\_mag and working\_sign respectively.
        \end{itemize}
    \end{itemize}
    \item After execution either:
    \begin{itemize}
        \item Take the $\text{working\_mag}[M] \times \text{working\_sign}[M]$.
        \item Optionally predict a selection $S$ to select the index $M$ in the working state.
    \end{itemize}
\end{itemize}
\textbf{Example Encoding of $(a+b) / (c*d)$}:
\begin{verbatim}
working mag
a,b,c,d
    O               G
0: [1,1,0,0,0,0,0]  [G = 1]
1: [0,0,1,1,0,0,0]  [G = 0]
2: [0,0,0,0,1,-1,0] [G = 1]
\end{verbatim}

\section{Future Work}
Two key research directions emerge from this work:

\begin{enumerate}
    \item \textbf{Integration with Large Language Models}: Document whether LLM performance on mathematics is improved by the inclusion of NAUs. This can be achieved by analyzing pretraining performance on math-heavy datasets like Proof-Pile-2 \cite{azerbayev2023llemma}, and the interpretability of NAUs should allow us to observe the operations being encoded from their respective input contexts.
    
    \item \textbf{Unified Arithmetic Operations}: Resolve the saddle point issues in internal operation space to allow a single NAU to extrapolate on all arithmetic operations, eliminating the need for operation-specific configurations like DMU\_add and DMU\_sub.
\end{enumerate}

\section*{Appendix}
\subsection*{DMU Convergence Thresholds}
The following table shows the convergence thresholds used to evaluate DMU performance across different operations and ranges. These thresholds were determined using the weight perturbation method described in Section 4.

\begin{table}[!htbp]
\centering
\caption{DMU Convergence Thresholds by Operation and Range}
\label{tab:dmu_thresholds}
\begin{tabular}{l|cccc}
\toprule
\textbf{Range} & \textbf{Addition} & \textbf{Subtraction} & \textbf{Multiplication} & \textbf{Division} \\
\midrule
sym & 7.55e-07 & 1.31e-07 & 1.27e-05 & 4.55e-08 \\
neg & 1.14e-06 & 9.44e-08 & 2.35e-05 & 6.59e-08 \\
pos & 3.68e-07 & 1.67e-07 & 2.04e-06 & 2.53e-08 \\
n10 & 1.13e-06 & 1.76e-07 & 1.83e-05 & 9.25e-08 \\
p01 & 2.61e-08 & 3.42e-08 & 4.31e-09 & 1.06e-07 \\
n01 & 3.64e-07 & 7.68e-09 & 6.13e-08 & 4.07e-07 \\
p11 & 2.63e-07 & 3.04e-07 & 1.39e-06 & 2.26e-08 \\
n20 & 1.96e-05 & 8.12e-06 & 1.99e-01 & 1.99e-08 \\
p20 & 2.73e-04 & 9.47e-06 & 6.67e-02 & 3.19e-08 \\
\midrule
\textbf{Mean} & 3.30e-05 & 2.06e-06 & 2.95e-02 & 9.07e-08 \\
\bottomrule
\end{tabular}
\end{table}

\section*{LLM Usage Statement}
LLMs were used extensively in the development of this project including ChatGPT 5.0, Gemini, and Claude. Coding agents were used extensively as well including the Cursor coding agent set to 'Auto' and Claude Code.

\end{document}